\documentclass[runningheads,a4paper]{llncs}

\usepackage{amssymb}
\usepackage{graphicx}

\usepackage{url}

\usepackage{wrapfig}
\usepackage{amsmath}
\usepackage{booktabs}
\usepackage{subfig}
\usepackage{color}
\usepackage{hyperref}

\begin{document}

\mainmatter  

\title{Multiple Landmark Detection using Multi-Agent Reinforcement Learning}
\titlerunning{Multiple Landmark Detection using MARL}


\author{Athanasios Vlontzos%
\and Amir Alansary \and Konstantinos Kamnitsas
   \and   Daniel Rueckert \and Bernhard Kainz}
\authorrunning{A. Vlontzos et al.}


\institute{BioMedIA, Computing Dept. Imperial College London}

\toctitle{Lecture Notes in Computer Science}
\tocauthor{Authors' Instructions}
\maketitle


\begin{abstract}
	The detection of anatomical landmarks is a vital step for medical image analysis and applications for diagnosis, interpretation and guidance. Manual  annotation of  landmarks  is a tedious process that requires domain-specific expertise and introduces inter-observer variability. This paper proposes a new detection approach for multiple landmarks based on multi-agent reinforcement learning.  Our hypothesis is that the position of all anatomical landmarks is interdependent and non-random within the human anatomy, thus finding one landmark can help to deduce the location of others. Using a Deep Q-Network (DQN) architecture 
	we construct an environment and agent with implicit inter-communication such that we can accommodate $K$ agents acting and learning simultaneously, while they attempt to detect $K$ different landmarks. During training the agents collaborate by sharing their accumulated knowledge for a collective gain. We compare our approach with state-of-the-art architectures and achieve significantly better accuracy by reducing the detection error by $50\%$, while requiring fewer computational resources and time to train compared to the na\"ive approach of training $K$ agents separately. Code and visualizations available: \url{https://github.com/thanosvlo/MARL-for-Anatomical-Landmark-Detection}
\end{abstract}
\section{Introduction}\label{sec_introduction}
The exact localization of anatomical landmarks in medical images is a crucial requirement for many clinical applications such as image registration and segmentation as well as computer-aided diagnosis and interventions. 
For example, for the planning of cardiac interventions it is necessary to identify standardized planes of the heart, \textit{e.g.} short-axis and $2/4$-chamber views \cite{alansary2018automatic}. 
It also plays a crucial role for prenatal fetal screening, where it is used to estimate biometric measurements like fetal growth rate to identify pathological development~\cite{rahmatullah2012image}. 
Moreover, the mid-sagittal plane, commonly used for brain image  registration and assessing anomalies, is identified based on landmarks such as the Anterior Commissure (AC) and Posterior Commissure (PC)~\cite{Alansary2018}. 
Manual annotation of landmarks is often a time consuming and tedious task that requires significant expertise about the anatomy and suffers from inter- and intra-observer errors. Automatic methods on the other hand can be challenging to design because of the large variability in the appearance and shape of different organs, varying image qualities and artefacts. Thus, there is a need for methods that can learn how to locate landmarks with highest accuracy and robustness; one promising approach is based on the use Reinforcement Learning (RL) algorithms~\cite{ghesu2016artificial,Alansary2018}.

\noindent\textbf{Contributions:}
This work presents a novel Multi-Agent Reinforcement Learning (MARL) approach for detecting multiple landmarks efficiently and simultaneously by sharing the agents' experience.
The main contributions can be summarized as: \textit{(i)} We introduce a novel formulation for the problem of multiple landmark detection in a MARL framework; 
\textit{(ii)} A novel collaborative deep Q-network (DQN) is proposed for training using implicit communication between the agents; \textit{(iii)} Extensive evaluations on different datasets and comparisons with recently published methods are provided (decision forests, Convolutional Neural Networks (CNNs), and single-agent RL).

\noindent{\textbf{Related Work}}
In the literature, automatic landmark detection approaches have adopted machine learning algorithms to learn combined appearance and image-based models, for example using regression forests \cite{oktay2017stratified} and statistical shape priors \cite{gauriau2015multi}. 
Zheng et al.~\cite{zheng20153d} proposed using two CNNs for landmark detection; the first network learns the search path by extracting candidate locations, and the second learns to recognize landmarks by classifying candidate image patches. Li et al.~\cite{li2018fast} presented a patch-based iterative CNN to detect individual or multiple landmarks simultaneously. 
Ghesu et al.~\cite{ghesu2016artificial} introduced a single deep RL agent to navigate in a 3D image towards a target landmark. The artificial agent learns to search and detect landmarks efficiently in an RL scenario. This search can be performed using fixed or multi-scale step strategies \cite{ghesu2017multi}.
Recently, Alansary et al.~\cite{Alansary2018} proposed the use of different Deep Q-Network (DQN) architectures for landmark detection with novel hierarchical action steps. 
The agent learns an optimal policy to navigate using sequential action steps in a 3D image (environment) from any starting point towards the target landmark. In \cite{Alansary2018} the reported experiments have shown that such an approach can achieve state-of-the-art results for the detection of multiple landmarks from different datasets and imaging modalities.
However, this approach was designed to learn a single agent for each landmark separately. In~\cite{Alansary2018} it has also been shown that performance of different strategies and architectures strongly depends on the anatomical location of the target landmark. Thus we hypothesize that sharing information while attempting simultaneous detection reduces the aforementioned dependency.

\begin{figure}
	\centering
	\includegraphics[width=\linewidth]{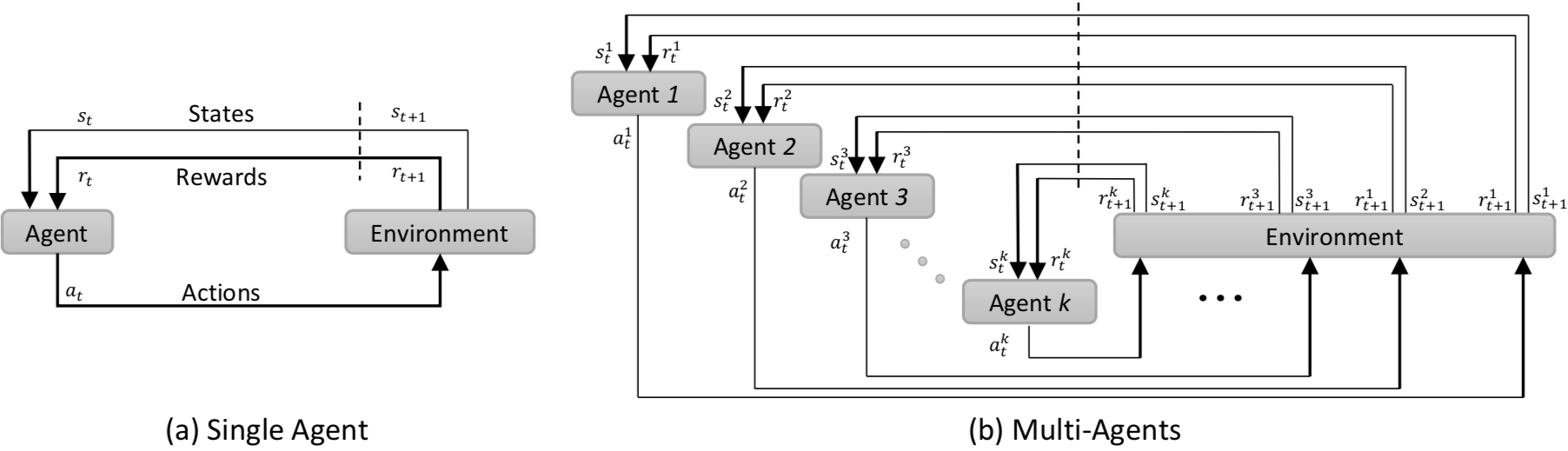}
	\caption{(a) A single agent and (b) multi agents interact within an RL environment.}
	\centering
	\label{fig_single_vs_multi_agents}
\end{figure}

\noindent\textbf{Background:}\label{sec_background}
Reinforcement Learning (RL) allows artificial agents to learn complex tasks by interacting with an environment $E$ using a set of actions $A$. The agent learns to take an action $a$ at every step (in a state $s$) towards the target solution guided by a reward signal $r$ during training. The main goal is to maximize the expected rewards in order to find the optimal policy $\pi^*$. 
In Q-Learning, a state-action value function $Q(s,a)$ is used to approximate the value of taking an action in a given state. 
The Q-function is defined as the expected value of the accumulated discounted future rewards, which can be approximated iteratively as: $Q_{t+1}(s,a)=E[r+\gamma {\max}_{a}(Q_{t}(s',a'))]$. Here $\gamma \in [0,1]$ is a discount factor that is used to incorporate the notion of uncertainty in future events. 
Mnih et al.~\cite{mnih2015human} proposed an approximation of the Q-function using a CNN by optimizing the network cost $L(\theta) = E \left[ \left( r + \gamma \: \underset{a'}{\max}\: Q_{target}(s',a';{\theta }^{-}) - Q_{net}(s,a;\theta) \right)^{2} \right]$. $Q_{target}$ is a temporary fixed version of $Q_{net}$, which gets updated every $N_{target}$ steps, used in order to avoid destabilization caused by rapid policy changes. \\
In single-agent RL scenarios, individual models learn solely from states that result from the actions of an agent. Complementary to this, MARL models learn from states that result from multiple agents dynamically interacting with their shared environment.
In MARL models, there are $K$ agents interacting with environment $E$. Each learns to take an action $a_t^k$ during a state $s_t^k$ using a reward signal $r_t^k$. Thus, the environment is subjected to the actions of all agents, as shown in Figure~\ref{fig_single_vs_multi_agents}. Hence, the environment becomes non-stationary as action $a_i$ in state $s_k$ will not always lead to the same future, since the future state is also a function of the other agents. This causes a violation of the Markov assumptions needed for the formulation of a RL scenario as a Markov Decision Process (MDP). To address this issue, \cite{foerster2017learning,rashid2018qmix} proposed to establish communication between the agents, thus taking all agents actions into account. \\
Any agent communication signifies the exchange of information or knowledge about the underlying Markov state of the environment. Communication between agents can be achieved explicitly via a communication protocol like in~\cite{Foerster2016}, where a limited bandwidth channel is learned by the agents, or implicitly by sharing knowledge in the parameter space or by combining value functions~\cite{Gupta2017}.
MARL scenarios can be classified as collaborative or competitive depending on the relation of the communication between agents. In this paper, we define the collaborative scenario as agents that attempt to minimize a common loss function. Competition between agents signifies a scenario in which agents try to minimize their own loss function through increasing the loss function of  other agents. 

\section{Proposed Method\label{method}}
In this work, we formulate the problem of multiple anatomical landmark detection as a multi-agent reinforcement learning scenario. Building upon the work of \cite{ghesu2016artificial} and \cite{Alansary2018} we extend the formulation of landmark detection as a Markov Decision Process (MDP), where artificial agents learn optimal policies towards their target landmarks, which defines a concurrent Partially Observable Markov Decision Process (co-POMDP)\cite{Girard2015}. 
We consider our framework concurrent as the agents train together but each learns its own individual policy, mapping its private observations to a personal action~\cite{Gupta2017}. We hypothesize that this is necessary as the localization of different landmarks requires learning partly heterogeneous policies. This would not be possible with the application of a centralized learning system.
 
Our RL framework is defined by the \emph{States} of the environment, the \emph{Actions} of the agent, their \emph{Reward Function} and the \emph{Terminal State}. 
We consider the environment to be a 3D scan of the human anatomy and define a state as a Region of Interest (ROI) centred around the location of the agent.
This makes our formulation a POMDP as the agents can only see a subset of the environment\cite{Jaakkola1995}. We define the frame history to be comprised of four ROIs. 
In this setup each agent can move along the $x,y,z$ axis creating thus a set of six actions. 
The agents evaluate their chosen actions based on the maximization of the rewards received from the environment. The reward function is defined as the relative improvement in Euclidean distance between their location at time $t$ and the target landmark location. In our multi-agent framework, each agent calculates its individual reward as their policies are disjoint. 

During training, we consider the search to have converged when the agent reaches a region within 1mm of the target landmark. Episodic play is introduced in both training and testing. In training, the episode is defined as the time the agents need to find the landmarks or until they have completed a predefined maximum number of steps. In case one agent finds its landmark before all others, we freeze the training and disable network updates derived from this agent while allowing the other agents to continue exploring the environment. During testing, we terminate the episode when the agent starts to oscillate around a position or exceeds a defined maximum number of frames seen in the episode similar to~\cite{Alansary2018}. 

\noindent{\textbf{Collaborative Agents}}
Previous approaches to the problem of landmark detection by \cite{Alansary2018}, \cite{ghesu2017multi} and \cite{ghesu2016artificial} considered a single agent looking for a single landmark. This means that further landmarks needs to be trained with separate instances of the agent making a large scale application unfeasible. 
Our hypothesis is that the position of all anatomical landmarks is interdependent and non-random within the human anatomy, thus finding one landmark can help to deduce the location of other landmarks. 
This knowledge is not exploited when using isolated agents. Thus, in order to reduce the computational load in locating multiple landmarks and increase accuracy through anatomical interdependence, we propose a collaborative multi agent landmark detection framework (Collab-DQN). 
The following description will assume just two agents for simplicity of presentation. However, our approach scales up to $K$ agents. For our experiments we show evaluations using two, three and five agents trained together. 

\begin{figure}
	\centering
	\includegraphics[width=.8\linewidth]{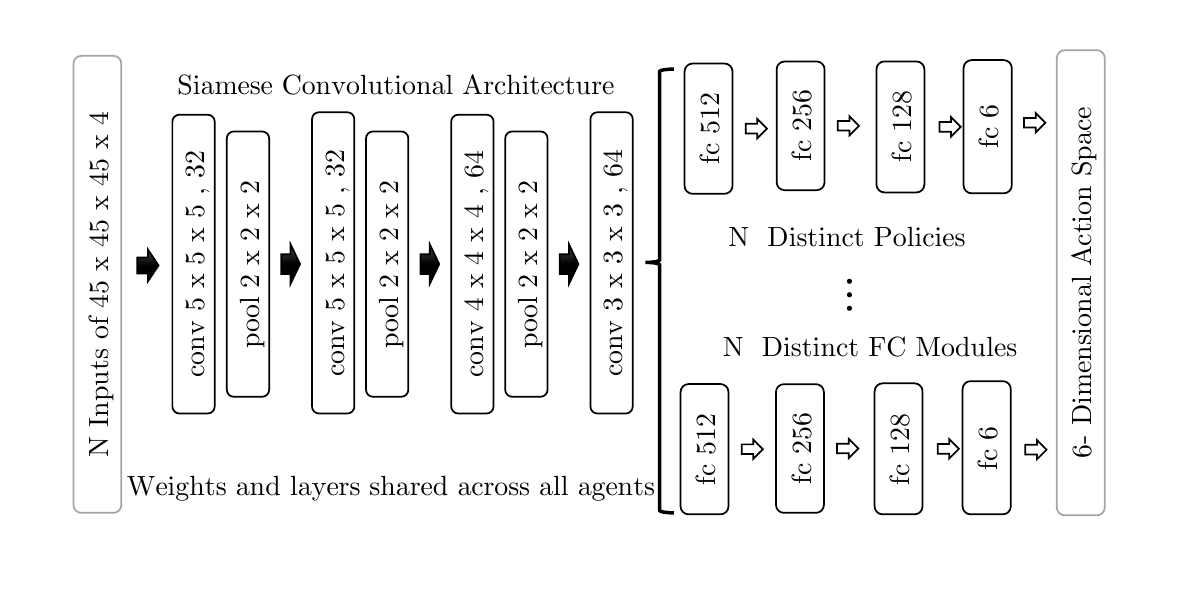}
	\caption{Proposed Collaborative DQN for the case of two agents; The {\tt{convolutional}} layers and corresponding weights are shared across all agents making them part of a Siamese architecture, while the policy making fully connected layers are separate for each agent}
	\label{colab_dqn_framework}
\end{figure}

A DQN is composed of three {\tt{convolutional}} layers interleaved with {\tt{maxpool}} layers followed by three {\tt{fully connected}} layers. Inspired by Siamese architectures~\cite{Bromley1994}, in our Collab-DQN we build $K$ DQN networks with the difference that weights are shared across the {\tt{convolutional}} layers. The {\tt{fully connected}} layers remain independent since these will make the ultimate action decisions constituting the policy for each agent. In this way, the information needed to navigate through the environment are encoded into the shared layers while landmark specific information remain in the fully connected ones. In Figure~\ref{colab_dqn_framework}, we graphically represent the proposed architecture for two agents. 
Sharing the weights across the {\tt{convolutional}} layers helps the network to learn more generalized features that can fit both inputs while adding an implicit regularization to the parameters avoiding overfitting. The shared weights enable indirect knowledge transfer in the parameter space between the agents, thus, we can consider this model as a special case of collaborative learning~\cite{Gupta2017}.

\section{Experimentation}
\noindent{\textbf{Dataset:}}
We evaluate our proposed framework and model on three tasks: (i) brain MRI landmark detection with 728 training and 104 testing volumes~\cite{jack2008alzheimer}; (ii) cardiac MRI landmark detection with 364 training and 91 testing volumes~\cite{de2014population} and (iii) landmark detection in fetal brain ultrasound with 51 training and 21 testing volumes. Each modality includes 7-14 anatomical ground truth landmark locations annotated by expert clinicians~\cite{Alansary2018}.\\
\noindent{\textbf{Training:}} During training an initial random location is chosen from the inner $80\%$ of the volume, in order to avoid sampling outside a meaningful area. The initial ROI is $45\times45\times45$ pixels around the randomly chosen point. The agents follow an $\epsilon$-greedy exploration strategy, where every few steps they choose a random action from a uniform distribution while during the remaining steps they act greedily. Episodic learning with the addition of freezing action updates for the agents that have reached their terminal state until the end of the episode is used, as detailed in Section~\ref{method}.

\noindent{\textbf{Testing:}}
For each agent, we fixed 19 different starting points in order to have a fair comparison among the different approaches. These points were used for all testing volumes for each modality at $25\%$, $50\%$ and $75\%$ of the volume's size. For each volume the Euclidean distance between the end location and the target location was averaged for each agent for each of the 19 runs. The mean distance in mm was considered to be the performance of the agent in the specific volume.
	
Multiple tests have been performed using our proposed architecture. Comparisons are made against the performance on multi-scale RL landmark detection~\cite{ghesu2017multi}, fully supervised deep Convolutional Neural Networks (CNN)~\cite{li2018fast} as well as a single agent DQN landmark detection algorithm~\cite{Alansary2018}. In case of cardiac landmarks we compare with~\cite{oktay2017stratified} that utilizes decision forests. Different DQN variations like the Double DQN or Duelling DQNs are not evaluated since their performance provides little to no improvement for the task of anatomical landmark detection as exhibited in~\cite{Alansary2018}.

	\begin{table}[t]
	\begin{center}
	\begin{tabular}{|l|l|l|l|l|l|}
		\hline
		\textbf{Method}               & \textbf{AC}             & \textbf{PC}                & \textbf{RC}            & \textbf{LC}              & \textbf{CSP}    \\ \hline
		
		\textbf{Supervised CNN}       & -                       & -                          & -                      & -                        & $5.47 \pm 4.23$ \\ \hline
		\textbf{DQN}                  & $2.46\pm1.44$           & $2.05\pm 1.14$                    & $3.37 \pm 1.54$        & $3.25\pm1.59$            & $\mathbf{3.66\pm 2.11}$  \\ \hline
		\textbf{Collab DQN}          & \textbf{$\mathbf{0.93\pm 0.18}$} & \textbf{$\mathbf{1.05 \pm 0.25}$}  & \textbf{$\mathbf{2.52\pm2.25}$} & \textbf{$\mathbf{2.41 \pm 1.52}$} &        $3.78\pm5.55$         \\ \hline
	\end{tabular}
	\caption{Results in millimeters for the various architectures on landmarks across brain MRI and fetal brain US. Our proposed Collab DQN performs better in all cases except the CSP where we match the performance of the single agent. }
	\label{results_total_table}
		\end{center}
\end{table}	
 Even though our method can scale up to $K$ agents given enough computational power we limited our comparison to the Anterior Commissure (AC) and the Posterior Commissure (PC) of the brain; the Apex (AP) and Mitral Valve Centre (MV) of the heart; the Right Cerebellum (RC), Left Cerebellum (LC) and Cavum Septum Pellucidum (CSP) for the fetal brain. 
 These are common, diagnostically valuable landmarks used in the clinical practice and by previous automatic landmark detection algorithms. 
 For completeness and to facilitate future comparisons, we provide our performance comparison also for the training of three and five agents simultaneously. In Table~\ref{results_total_table}, we show the performance of the brain MRI and fetal brain US landmarks using the different approaches. In Table~\ref{multi-landmark} we exhibit the results for three and five agents trained simultaneously and the results for cardiac MRI landmarks.

\begin{table}[b]
	\subfloat[]{
	\centering
			\begin{tabular}{|c|c|c| } 
				\hline
				\textbf{Landmark}       & \textbf{3 Agents} & \textbf{5 Agents} \\ \hline
				\textbf{AC}             & $0.94\pm0.17$ & $0.98\pm0.25$ \\
				\hline
				\textbf{PC}             & $0.96\pm0.20$ & $0.90\pm0.18$ \\ \hline
				\textbf{Landmark 3}     & $1.45\pm0.51$ & $1.39\pm0.45$ \\ \hline
				\textbf{Landmark 4}     & N/A           & $1.42\pm0.90$ \\ \hline
				\textbf{Landmark 5}     & N/A           & $1.72\pm0.61$ \\ \hline
		\end{tabular} 
		}
	\subfloat[]{
			\centering
		\begin{tabular}{|c|c|c| } 
				\hline
				\textbf{Method}                 & \textbf{AP}               & \textbf{MV} \\ \hline
				\textbf{Inter-Obs. Error}   & $5.79 \pm 3.28$           & $5.30 \pm 2.98$ \\
				\hline
				\textbf{Decision Forest}        & $6.74 \pm 4.12$           & $6.32\pm3.95$ \\ \hline
				\textbf{DQN}                    & $4.47 \pm 2.64$           & $5.73\pm4.16$ \\ \hline
				\textbf{DQN Batch $\times 2$}     & $4.30 \pm 12.07$           & $5.01 \pm 4.49$ \\ \hline
				\textbf{DQN Iterations $\times 2$}     & $4.78 \pm 13.87$           & $5.70 \pm 18.11 $ \\ \hline
				\textbf{Collab DQN}             & \textbf{$\mathbf{3.96 \pm 5.07}$}  & \textbf{$\mathbf{4.87 \pm 0.26}$} \\ \hline
		\end{tabular} 
	}
	\caption{(a)Multiple agent performance, training and testing were conducted in the Brain MRI; Landmarks 3,4,5 represent respectively the outer aspect, the inferior tip and the inner aspect of the splenium of corpus callosum; (b) multi-agent performance on cardiac MRI dataset; }\label{multi-landmark}
	\label{tab:results}
\end{table} 
\noindent \textbf{Discussion:} As shown in  Tables~\ref{results_total_table} and~\ref{tab:results} our proposed method significantly outperforms the current state-of-the-art in landmark detection. $p$-values from a paried student-t test for all experiments were in the range 0.01 to 0.0001. 
We perform an ablation study by training instances of a single agent with double the iterations and double the batch size. The study has been conducted on the Cardiac MRI landmarks that have exhibited the biggest localization difficulties because of larger anatomical variations across subjects than observed in brain data. 
Our results confirm that the agents share basic information across them, which helps all of them perform their tasks more efficiently. These results support our hypothesis that the regularization effect from the gradients collected from the increased experience and knowledge of the multi-agent system is advantageous. Furthermore, we created a single agent with doubled memory but due to the random initialization of experience memory, the agent failed to learn. In addition, as shown in Table~\ref{multi-landmark}(a), the inclusion of more agents leads to similar or improved results across all landmarks.
It is interesting to note that even though we perform better in all landmarks, our approach can only match the performance of a single agent DQN for the CSP landmark. We theorize that this is due to the different anatomical nature of the RC, LC landmarks compared to the CSP landmark, thus the joint detection does not present an advantage. We chose to utilize the DQN in this paper rather than existing policy gradient methods like A3C as the DQN is  represented by a single deep CNN that interacts with a single environment. A3C use many instances of the agent that interact asynchronously and in parallel. Multiple A3C agents with multiple incarnations of such environments are computationally expensive. In future work, we will investigate the application of other methods for multiple-landmarks detection using either collaborative or competitive agents.

\noindent{\textbf{Computational Performance: }}
Training multiple agents together does not only provide benefits in performance of landmark localization, it also reduces the time and memory requirements of training. Sharing the weights between the convolutional layers helps to reduce the trainable parameters by $5\%$ in case of two agents and by $6\%$ in case of three agents when compared with the parameters of two and three separate networks respectively. Furthermore, the addition of a single agent to our architecture reduces the required number of parameters by $6\%$ compared to a single standalone agent. Due to the regularization effect that multiple agents have on their training and the implicit knowledge transfer, the training time our approach needs on average 25.000-50.000 less time steps to converge compared with a single DQN and each training epoch needs approximately 30 minutes less than the training of 2 epochs in a separate single DQN (NVIDIA Titan-X, 12 GB). Inference is on par with a single agent at $\sim$20fps. 

\section{Conclusion} 
In this paper we formulated the problem of multiple anatomical landmark detection as a multi-agent reinforcement learning scenario, we also introduced Collab-DQN, a Collaborative DQN for landmark detection in brain and cardiac MRI volumes and 3D US. We train $K$ agents together looking for $K$ landmarks. The agents share their convolutional layer weights. In this fashion we exploit the knowledge transferred by each agent to teach the other agents. We achieve significantly better performance than the next best method of~\cite{Alansary2018} decreasing the error by more than 1mm while taking less time to train and less memory than training $K$ agents serially. We believe that a Bayesian exploration approach is a natural next step, which will be addressed in future work.

\noindent\textbf{Acknowledgements:} Wellcome Trust IEH Award [102431], EPSRC  EP/ S013687/ 1, NVIDIA for their GPU donations. Brain MRI: \url{adni.loni.usc.edu}, US data: access only with informed consent, subject to approval and  formal Data Sharing Agreement. Caridac data: \url{digital-heart.org}.


\bibliographystyle{splncs03}
\bibliography{references}

\end{document}